\documentclass[11pt, letterpaper]{nyu}
\usepackage[all]{hypcap}

\usepackage{hyperref}[citecolor=lightblue]

\hypersetup{
    colorlinks = true,
}

\usepackage{microtype}
\usepackage{graphicx}
\usepackage{subfigure}
\usepackage{booktabs} %
\usepackage{float}
\usepackage{amsmath}
\usepackage{amssymb}
\usepackage{mathtools}
\usepackage{amsthm}
\usepackage{caption}
\usepackage{mathrsfs}
\usepackage{nicefrac}
\usepackage{dsfont}
\usepackage{enumitem}
\usepackage{float}
\usepackage[authoryear, sort&compress, round]{natbib}
\usepackage{xspace}
\usepackage[capitalize,noabbrev]{cleveref}
\usepackage{subcaption}
\usepackage{wrapfig}
\usepackage{lipsum}
\usepackage{listings}
\usepackage{amsmath}
\usepackage{amssymb}
\usepackage{mathtools}
\usepackage{amsthm}
\usepackage{bbm}
\usepackage{algpseudocode}
\usepackage{setspace}
\usepackage{color}
\usepackage[skins,theorems]{tcolorbox}
\usepackage{xcolor}
\usepackage{tikz}

\definecolor{deepblue}{rgb}{0,0,0.5}
\definecolor{deepred}{rgb}{0.6,0,0}
\definecolor{deepgreen}{rgb}{0,0.5,0}

\newcommand\pythonstyle{\lstset{
basicstyle=\ttfamily\footnotesize,
language=Python,
morekeywords={self, clip, exp, mse_loss, uniform_sample, concatenate, logsumexp},              %
keywordstyle=\color{deepblue},
emph={MyClass,__init__},          %
emphstyle=\color{deepred},    %
stringstyle=\color{deepgreen},
frame=single,                         %
showstringspaces=false
}}

\lstnewenvironment{python}[1][]
{
\pythonstyle
\lstset{#1}
}
{}

\newcommand\pythoninline[1]{{\pythonstyle\lstinline!#1!}}

\makeatletter
\def\mathcolor#1#{\@mathcolor{#1}}
\def\@mathcolor#1#2#3{%
  \protect\leavevmode
  \begingroup
    \color#1{#2}#3%
  \endgroup
}
\makeatother

\usepackage[textsize=tiny]{todonotes}

\usepackage{crossreftools}
\usepackage{dsfont}
\usepackage{nicefrac}
\usepackage{inconsolata}
\usepackage{algorithm}
\usepackage{amssymb}
\usepackage{csquotes}

\newcommand{\xxnote}[3]{}
\renewcommand{\xxnote}[3]{}

\newcommand{\website}{\href{https://robotutilitymodels.com}{robotutilitymodels.com}}

\newcommand{\longname}{Robot Utility Models}
\newcommand{\longnamesingular}{Robot Utility Model}
\newcommand{\shortname}{RUMs}
\newcommand{\shortnamesingular}{RUM}
\newcommand{\finalSuccessRate}{90\%}

\title{Robot Utility Models: General Policies for Zero-Shot Deployment in New Environments}
\websitedef{\website}

\reportnumber{} 

\author[1]{Haritheja Etukuru$^{*}$}
\author[1]{Norihito Naka}
\author[1]{Zijin Hu}
\author[1]{Seungjae Lee}
\author[2]{Julian Mehu}
\author[2]{Aaron Edsinger}
\author[2]{Chris Paxton}
\author[3]{Soumith Chintala}
\author[1]{Lerrel Pinto}
\author[1,2]{Nur Muhammad ``Mahi'' Shafiullah$^{*}$}

\affil[1]{New York University}
\affil[2]{Hello Robot Inc.}
\affil[3]{Meta Inc.}

\correspondingauthor{ \href{mailto:mahi@cs.nyu.edu}{mahi@cs.nyu.edu}. $(^{*})$~denotes equal contribution.}

\begin{abstract}
\vspace{-1em}
Robot models, particularly those trained with large amounts of data, have recently shown a plethora of real-world manipulation and navigation capabilities. Several independent efforts have shown that given sufficient training data in an environment, robot policies can generalize to demonstrated variations in that environment. However, needing to finetune robot models to every new environment stands in stark contrast to models in language or vision that can be deployed zero-shot for open-world problems.
In this work, we present \textit{\longname{} (\shortname{})}, a framework for training and deploying zero-shot robot policies that can directly generalize to new environments without any finetuning. To create \shortname{} efficiently, we develop new tools to quickly collect data for mobile manipulation tasks, integrate such data into a policy with multi-modal imitation learning, and deploy policies on-device on Hello Robot Stretch, a cheap commodity robot, with an external mLLM verifier for retrying. We train five such utility models for opening cabinet doors, opening drawers, picking up napkins, picking up paper bags, and reorienting fallen objects. Our system, on average, achieves \textbf{\finalSuccessRate{} success rate in unseen, novel environments interacting with unseen objects}. Moreover, the utility models can also succeed in different robot and camera set-ups with no further data, training, or fine-tuning. 
Primary among our lessons are the importance of training data over training algorithm and policy class, guidance about data scaling, necessity for diverse yet high-quality demonstrations, and a recipe for robot introspection and retrying to improve performance on individual environments.
\end{abstract}

\begin{document}
\maketitle

\begin{figure}[h]
    \vskip -0.5em
    \centering
    \includegraphics[width=1.0\linewidth]{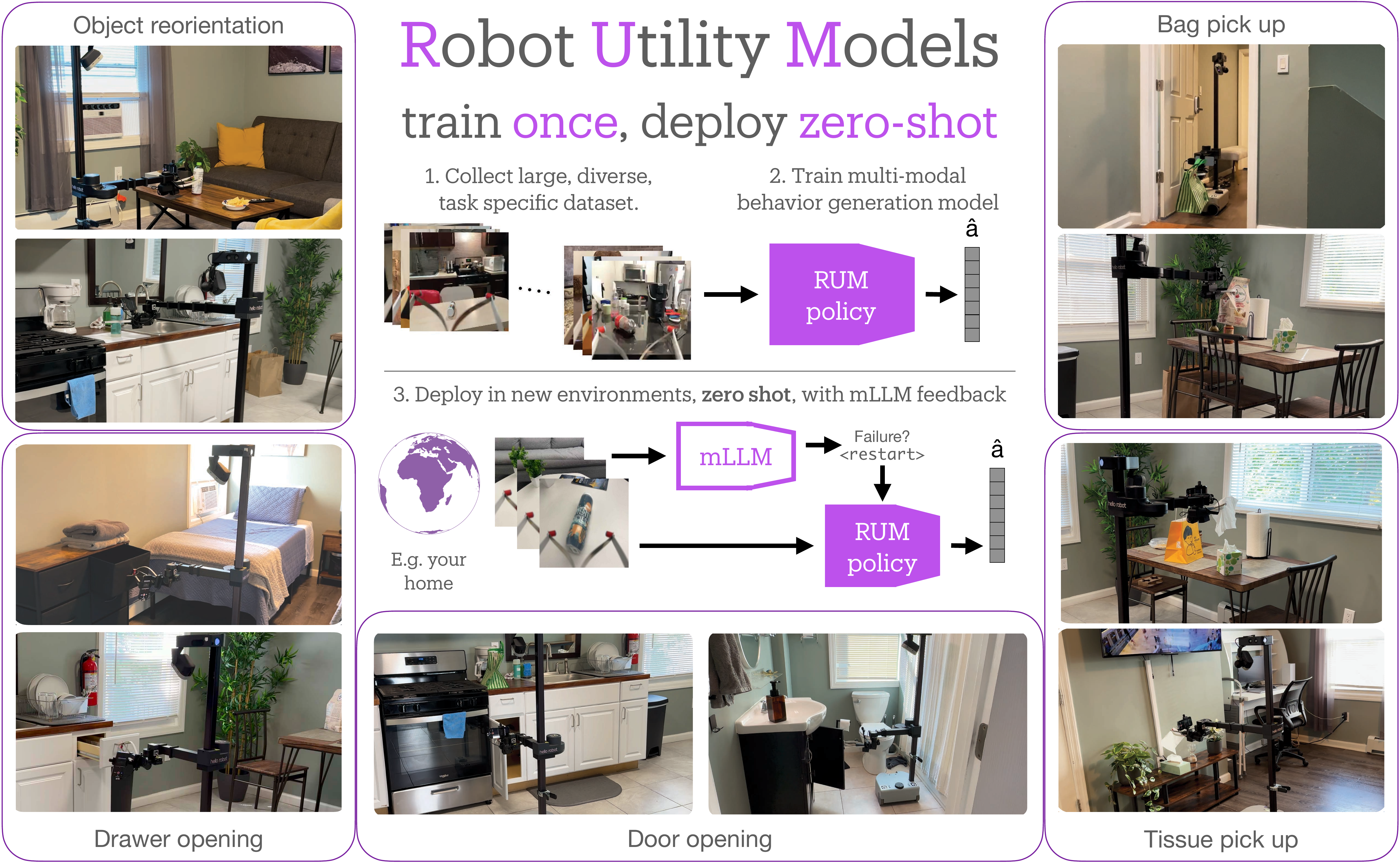}
    \caption{\longname{} are trained on a diverse set of environments and objects, and then can be deployed in novel environments with novel objects without any further data or training.}
    \label{fig:intro-figure}
    \vskip -1em
\end{figure}

\section{Introduction}

We have seen rapid progress in training manipulation skills recently~\citep{zhao2023aloha,fu2024mobile,brohan2023rt2,haldar2024baku,fu2024humanplus,lin2024learning,kim2024openvla}, largely brought about by fitting deep networks on data collected by teleoperating robots~\citep{mandlekar2018roboturk,iyer2024open,arunachalam2023dexterous,cheng2024opentelevisionteleoperationimmersiveactive,khazatsky2024droid}. The mechanism for deploying such skills in new environments mimics the pretrain-then-finetune strategy first developed by the vision community circa 2014~\citep{girshick2014rich}. There, models were first pretrained on ImageNet and then finetuned on task-specific data such as detection, segmentation, and pose estimation~\citep{girshick2014rich,GkioxariHGM14}.
In the context of robotics, this strategy involves pretraining on large robot datasets~\citep{padalkar2023open,khazatsky2024droid,shafiullah2023bringing,walke2023bridgedata} to produce a robot foundation model, which is then fine-tuned on data collected in new environments or tasks~\citep{shafiullah2023bringing,team2024octo,kim2024openvla}. This need to fine-tune the foundation model for each and every new environment is limiting as it requires humans to collect data in the very environment where the robot is expected to perform. So while vision and language models have moved on to zero-shot deployments, i.e. without any environment-specific finetuning data, such a capability eludes most robot manipulators. This is not to say that there have not been attempts to create zero-shot manipulation models -- several foundational work in grasping and pick-and-place~\citep{fang2023anygrasp,sundermeyer2021contact,Mahler2017DexNet2D} have tackled this problem albeit with a task-specific solution.

So what makes creating a general policy for an arbitrary task that can work zero-shot hard? First is the concern about sufficient data -- the necessary amount of data to train such a general model could be large. Since collecting robot data is hard, creating a large dataset is also hard and often expensive since humans are usually tasked to collect robot demonstrations. Second, when a large dataset is collected in the open-world it would necessarily have large diversity and multiple modes in demonstrator behavior. Fitting a robot models on this diverse data is a challenge. Third, unlike vision and language, where the native form of data, i.e. images and text are largely standard, robotics is far from having a standard camera and hardware setup along with physical challenges of running models in realtime on onboard compute. Creating zero-shot models that can run with even minor changes to hardware setup between training and deployment requires careful attention to details.
Finally, any model deployed zero shot on a novel environment naturally has a higher failure rate than a model that has been fine-tuned on that environment. Thus, to deploy a model zero-shot, it is important to have a mechanism for error detection and recovery.

In this work, we introduce~\longname{} (\shortname{}), a new framework for training focused and functional \textit{utility} models to complete helpful tasks that can be deployed zero-shot \textbf{without further training or fine-tuning} in novel environments. This is done by taking a systems-first approach. To scale up our datasets without compromising on data quality, we develop a new tool, building on prior work in untethered data collection~\citep{shafiullah2023bringing,chi2024universal}. We train policies on these diverse dataset with state-of-the-art multi-modal behavior learning algorithms~\citep{lee2024behavior,chi2023diffusion} and show how they can absorb and scale with large-scale demonstration data. Finally, we deploy the policy in multiple different environments out of the box, with self-critique via mLLMs~\citep{guo2023doremi} and retrying, showing how the policy can be robustly executed on cheap, general-purpose hardware. A selection of our trained models are available on the Hello Robot Stretch without much modifications. Beyond the default Stretch deployment, we also enable deployment on other robot arms, cameras, and lighting conditions, showing the generalizability of our approach.

Creating and deploying~\shortname{} led us to several interesting lessons. First, we find that the quantity and quality of data is crucial for training a utility model, with the choice of model architecture being less critical. Second, we see that the diversity of the data collected is crucial for the model to generalize to new environments, and more important than the raw quantity of data. Third, we find that the model can be made more capable in single environments by performing self-critique on the model performance with an independent model and retrying when appropriate.

To validate \shortname{}, we run a total of 2,950 robot rollouts in real-world environments including homes in New York City (NY), Jersey City (NJ), and Pittsburgh (PA). These experiments reveal the following:
\begin{itemize}[leftmargin=*]
    \item We show that it is possible to create general \longname{} with a moderate amount of data in the  order of 1,000 demonstrations (Section~\ref{sec:methods}). These \shortname{} achieve a 90\% average success rate on zero-shot deployment in 25 novel environments (Section~\ref{sec:unseen-environments}).
    \item The success of \shortname{} relies primarily on two key techniques. First, the use of multi-modal policies (Section~\ref{sec:model-training}) provides a zero-shot success rate of 74.4\% (Section~\ref{sec:policy-architecture}). Second, the mLLM based self-critique and retrying system (Section~\ref{sec:retrying}) further improves the success rate by 15.6\% (Section~\ref{sec:introspection-retrying}).
    \item While the overall framework for \shortname{} is straightforward, the devil is in the details, where we find gains from unexpected sources, e.g. data diversity vs. data quantity (Section~\ref{sec:data-diversity} and~\ref{sec:expert-vs-play}).
\end{itemize}

To encourage the development of \shortname{} for a wider variety of tasks, our code, data, models, hardware designs, as well as our experiment and deployment videos are open sourced and can be found on our website: \website.
\section{Robot Utility Models}
\label{sec:methods}
We take a full-stack approach to create \longname{}. At its core, our system follows the imitation learning framework.
However, to effectively scale imitation learning to the point where our trained policies are deployable zero-shot, we create new tools and techniques to improve data collection, model training, inference, and deployment.

\subsection{Data collection tool}
\label{sec:data-collection-tool}
One of the primary requirements of our system is to be able to scale up diverse yet accurate demonstration data for cheap. To this end, we continue on the evolutionary path of hand-held, portable data collection tools~\citep{song2020grasping,young2020visual,pari2021surprising,shafiullah2023bringing,chi2024universal} that let us quickly collect precise demonstrations. Following our previous work~\citep{shafiullah2023bringing}, we call this tool \textit{Stick-v2}, which is a hand-held data collection tool built out of an iPhone Pro and a bill of materials that adds up to \$25. We combine inspirations from the quick deployability of Stick-v1, and the compact, handheld form factor of UMI gripper. For a detailed build instruction and the bill of materials, we refer the reader to the supplementary materials (Appendix~\ref{sec:app:bom}).

Our design decisions are predicated on a few factors: portability, convenience, and set-up speed. We experimentally found these factors to be important to quickly scale up robot datasets and training \shortname{}. As we show with experiments in Section~\ref{sec:scaling-datasets}, one of the most crucial aspect of data collection for \shortname{} is data diversity, i.e. collecting data from a large number of diverse environmets. Thus, it is crucial to have a portable tool that is easy to mass-print, carry, and deploy in a new environment. Secondly, it is important for the collected data to be accurate across many environments with many variations. Finally, it is important to minimize the ``per-environment set-up time'', whether that time is spent setting up the data collection system, calibrating the camera, or the tool's SLAM system.

\begin{figure}[t!]
    \centering
    \includegraphics[width=\textwidth]{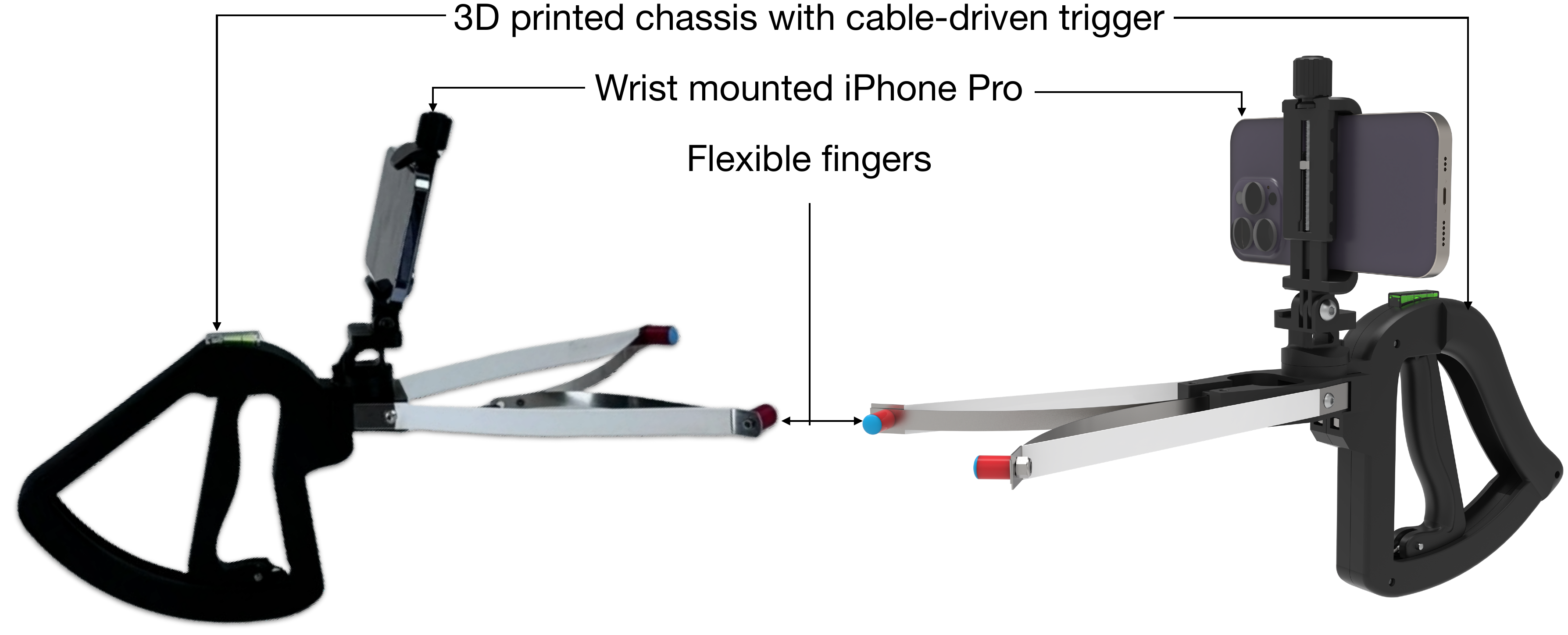}
    \caption{Stick-v2, our data collection tool (left: real photo, right: render), is built out of an iPhone Pro and a bill of materials that adds up to \$25. The tool is portable, robust, and makes it easy to start collecting data in a new environment in seconds.}
    \label{fig:stick-v2}
\end{figure}
For the above reason, we design our data collection tool, Stick-v2, around the ARKit API from the widely available and used iPhone Pro (Figure~\ref{fig:stick-v2}). Given its technical capabilities, the only digital component in our Stick-v2 is this iPhone, which makes our tool particularly robust to shipping and handling. The iPhone, and therefore Stick-v2, can collect RGB video and depth data at up to 60 Hz and high precision 6D pose and position information from the ARKit API at up to 100Hz. To capture the gripper opening information, we trained an RGB-based model that predicts the gripper aperture from images. Furthermore, this data is automatically synchronized and timestamped by the iPhone without the need for any calibration. This allows us to collect data from a wide variety of environments with no set-up time. This is in contrast to other data collection tools based on visual SLAM systems which has limited precision and are non-robust around ``textureless'' scenes such as close to flat walls, ceilings, or corners~\citep{chi2024universal,young2020visual}. Finally, not needing camera calibration makes our system deployable out-of-the-box in any environment, especially in the real world where the environment is not controlled. This enables us to, for example, collect data from retail home goods stores with minimal interruptions to enrich our datasets, which would be hindered if we had to calibrate the camera and odometry system for each new environment.

\subsection{Collected datasets}
\label{sec:collected-datasets}
We collect data for each of our five tasks, which are as defined below:
\begin{itemize}[leftmargin=*]
    \item \textbf{Door opening:} Open doors with a long handle, on e.g. cabinets and microwaves. Due to hardware limitations, our robot cannot open doors with round knobs, so we exclude them from our dataset.
    \item \textbf{Drawer opening:} Open a drawer with a handle. We exclude drawers with knobs from our dataset for similar reasons as above.
    \item \textbf{Reorientation:} Pick up a cylindrical object (e.g. bottle) lying on a flat surface and place it upright on the same surface.
    \item \textbf{Tissue pickup:} Pick up a soft, flexible tissue paper from any tissue paper box.
    \item \textbf{Bag pickup:} Pick up a kraft paper bag or similar other bags from a flat surface.
\end{itemize}
\begin{figure}[t!]
    \centering
    \includegraphics[width=\textwidth]{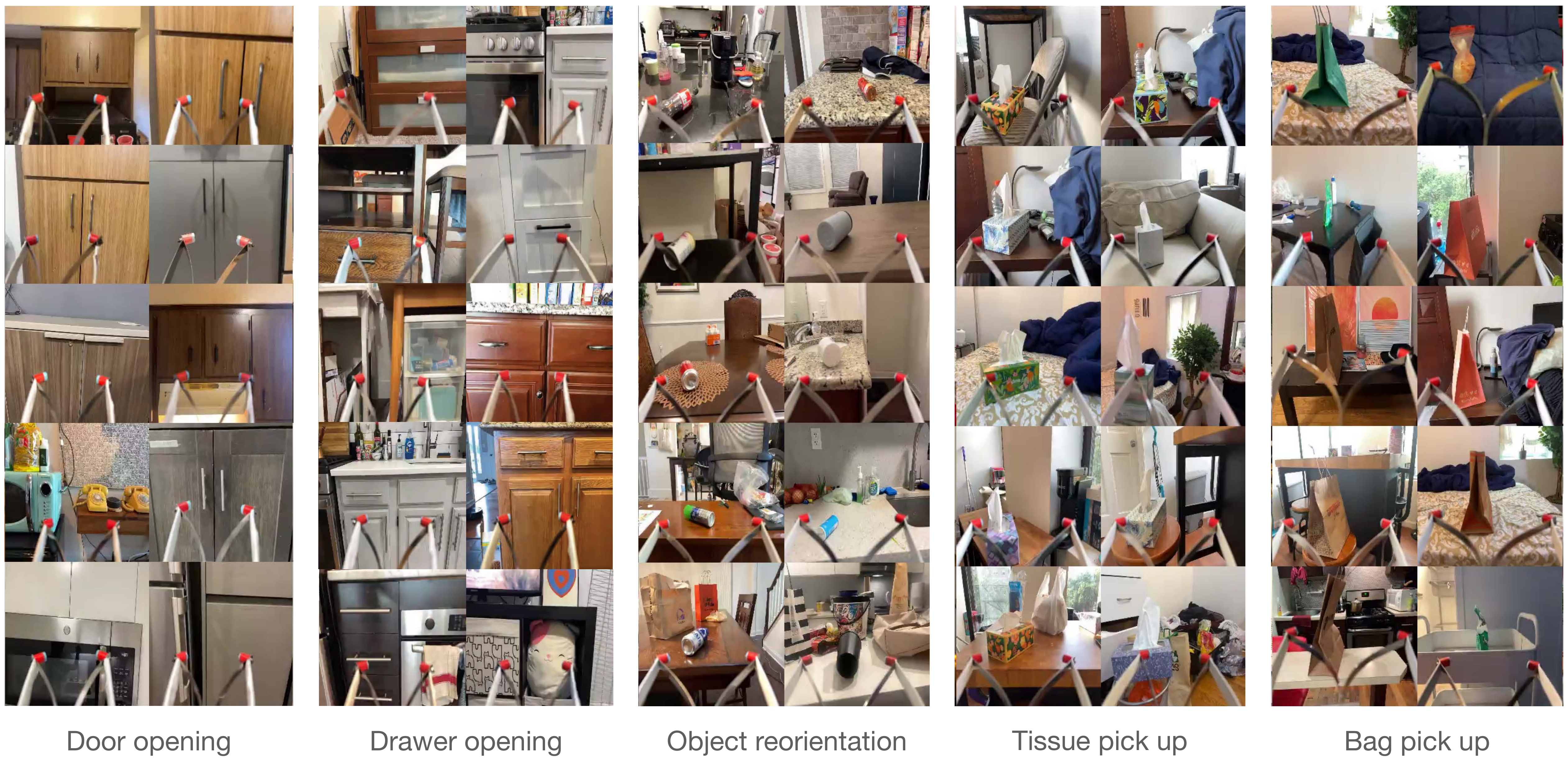}
    \caption{A small sample of environment and objects from our collected dataset. We collect data for each of our five tasks on a diverse set of environments and objects using Stick-v2.}
    \label{fig:dataset-fig}
\end{figure}
For each of our five~\shortname{}, we focused on gathering approximately 1,000 demonstrations
on approximately~40 environments, with about 25 demonstrations per environment on average. The only exceptions are door opening with~1,200 and drawer opening with~525 demonstrations. A small collection of such environments are shown in Figure~\ref{fig:dataset-fig}. For the door opening task, we seeded this dataset with the Homes of New York dataset~\citep{shafiullah2023bringing} as well as demonstrations collected during the Dobb$\cdot$E experiments. For the other tasks, our dataset consists of new demonstrations collected using the Stick-v2 tool on a novel set of environments and objects. For demonstrations collected from the previous dataset by inexperienced data collectors, we do a manual quality check and exclude any environment that has a high number of low-quality demonstrations, such as failed demonstrations. 
Note that, to keep our experiments unbiased, we hold out test environments and objects and never collect any data on them. To gain quick insight on different task data we use for training, we created an interactive data diversity visualization tool: \href{https://robotutilitymodels.com/data_diversity/}{robotutilitymodels.com/data\_diversity/}.

\subsection{Model training}
\label{sec:model-training}
Given that our data is collected by a large set of demonstration collectors, conceptually it is important for the model to handle any resultant multi-modality in the dataset. In this work, we train a large set of policy classes on our datasets for each task. Among the policy classes, the best performing ones are VQ-BeT~\citep{lee2024behavior} and Diffusion Policy (DP)~\citep{chi2023diffusion}. We also train ACT~\citep{zhao2023aloha} and MLP-BC policies on a limited set of tasks. Each policy class shares some features, such as a ResNet34-based vision encoder initialized to the HPR encoder from~\cite{shafiullah2023bringing}, and a transformer-based policy trunk. We also train each model for the same 500 epochs. Beyond that, we sweep to find the best hyperparameters for learning rate, history length, and chunk size, and use the recommended hyperparameters from the original papers for each model.
Our final VQ-BeT models are trained on data subsampled at 3.75Hz, and uses 6 most recent frames of history to predict the next action. All of our models predict the action in relative 6D space for the robot end-effector, and absolute value in the range $[0, 1]$ for the gripper opening.
We discuss the impact of choosing different training algorithms in Section~\ref{sec:policy-architecture}.
Training all of our models took between 24 and 48 hours on 2 Nvidia A100 GPUs on our cluster, with proportional speed-ups by using more GPUs or using more recent GPUs like H100s.

\subsection{Retrying with GPT-4o feedback}
\label{sec:retrying}
\begin{figure}[ht!]
    \centering
    \includegraphics[width=\textwidth]{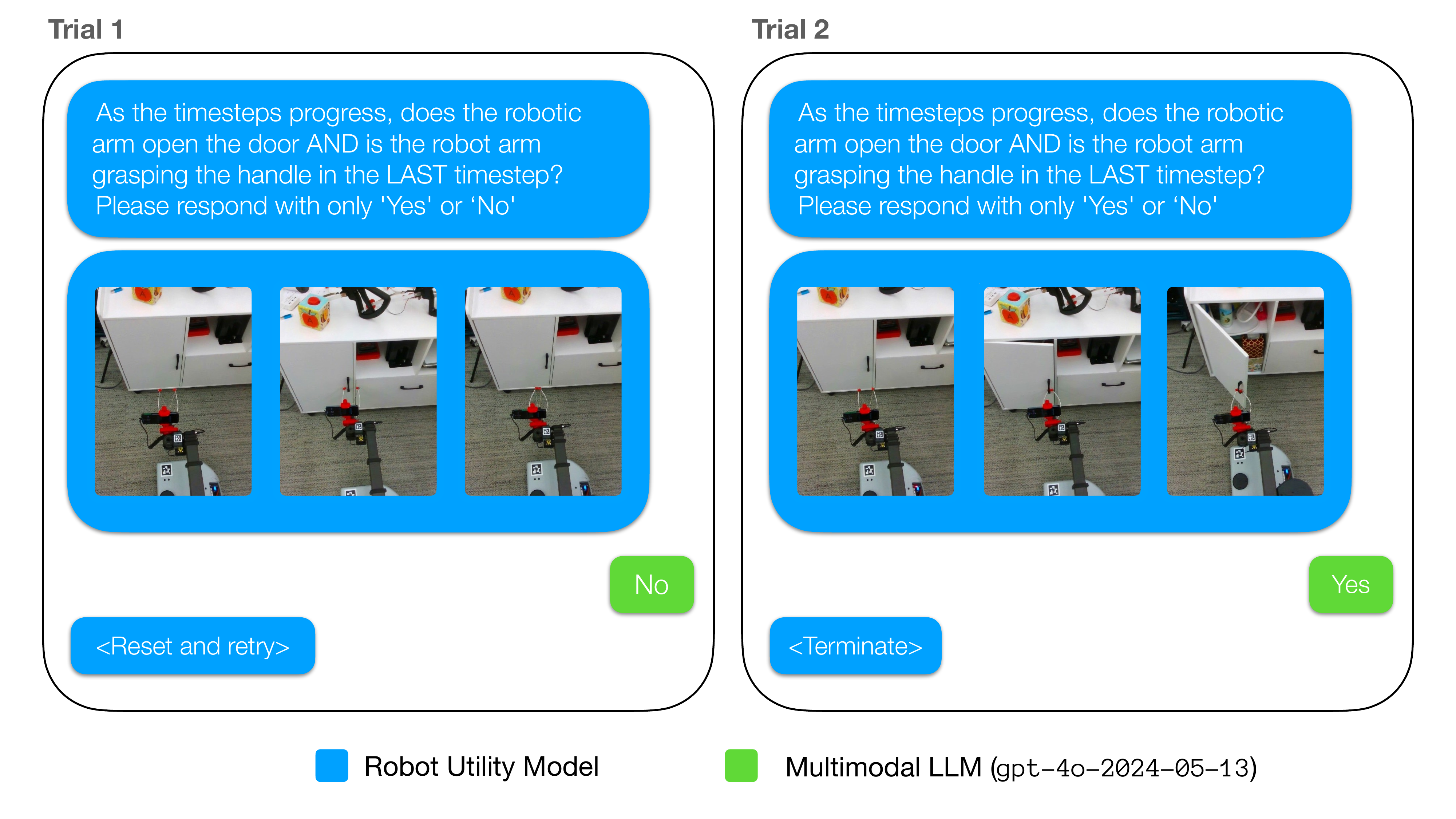}
    \caption{Automated retrying with feedback from multimodal LLM critic. We use a multimodal LLM (\texttt{gpt-4o-2024-05-13} in our experiments) to verify the success of a task given a summary of robot observations. If the mLLM detects a failure, we automatically reset the robot and retry the task with a new initial robot state until success or timeout.}
    \label{fig:mllm-feedback}
\end{figure}

While a pre-trained model can solve the task in a new environment, to achieve the best possible performance, it is helpful to have additional runtime support for the model. For our deployment, we use an multimodal LLM (\texttt{gpt-4o-2024-05-13}) as an introspection module for our policies for a success detection and retrying mechanism. We define a single verification prompt for each task, and ask the mLLM to verify the success of the task given a summary of robot observations. As for the run summary, we give the mLLM every other frame from the robot camera, which is either from the head or the wrist camera depending on the task. If the mLLM detects a failure (Figure~\ref{fig:mllm-feedback}), ~\shortnamesingular{} automatically resets the robot to a home position and retries the task with a new initial robot state.

\subsection{Deployment Details}
\label{sec:deployment}
\begin{figure}[t!]
    \centering
    \includegraphics[width=\textwidth]{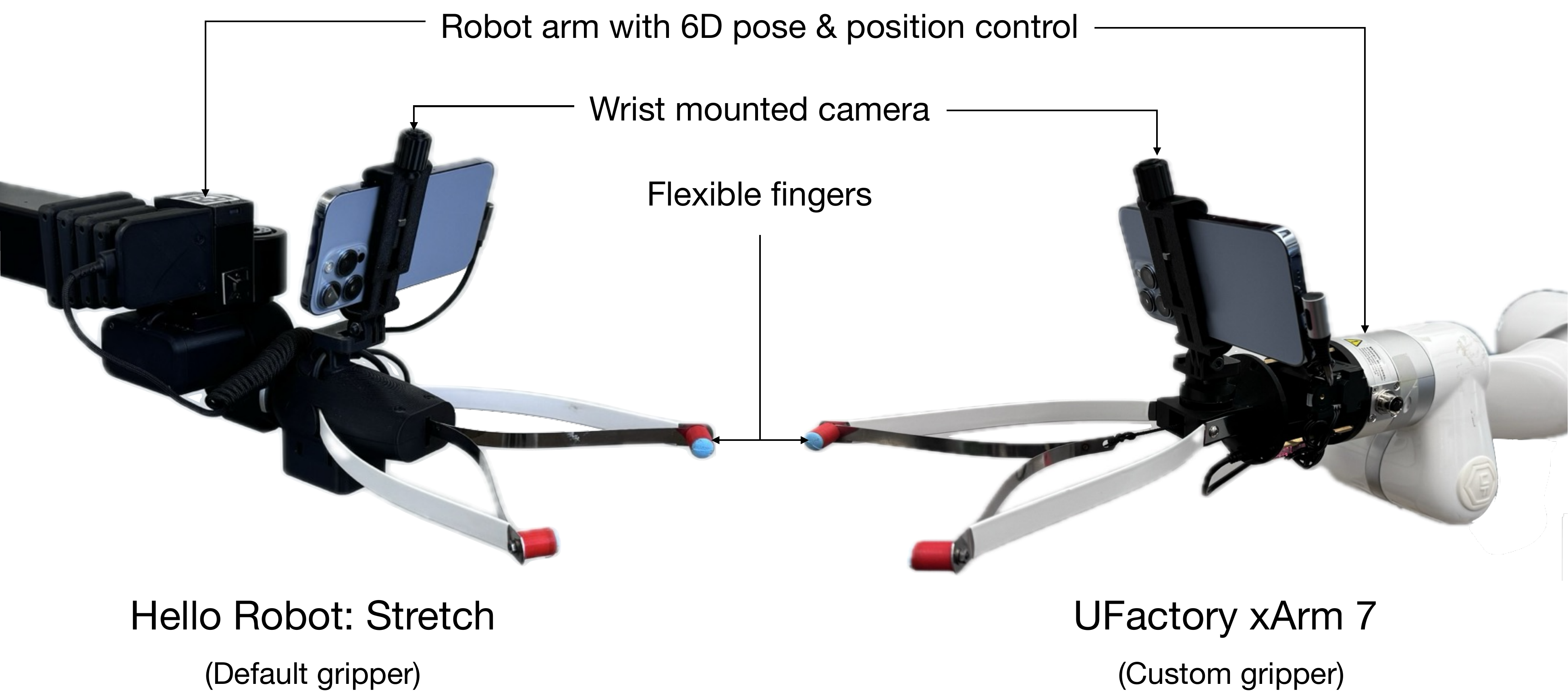}
    \caption{Picture of the some robot setups where our~\longname{} can be deployed. We show the Hello Robot: Stretch, and the xArm 7 robot with iPhone Pros on the wrist. Beyond these, we also deploy on Stretch robots with default D405 wrist cameras.
    }
    \label{fig:robot-setup}
\end{figure}
Our primary hardware for \longname{} deployment is the Hello Robot: Stretch robots with an iPhone on the wrist, but we support deploying our models on any robot arm with relative 6D pose and position control (Figure~\ref{fig:robot-setup}). We design and release an associated robot end-effector that can be mounted on standard robot arms, such as the xArm or Franka Panda. Similarly, while we primarily use the iPhone Pro as the deployment camera, we also show deployment on other wrist cameras, such as the Intel Realsense D405, which is the default wrist camera on Hello Stretch Edition 3 onwards. Overall, our deployment hardware system really relies on three things: our end-effector with a flexible two-fingered gripper and gripper tips, a wrist camera with a sufficient field of view, and an arm with six degrees of freedom to mount our wrist. We release default integration code for Hello Stretch 3 and an xArm wrist mount that we created, which should serve as illustrative examples for other arms.

\section{Capabilities of Robot Utility Models}
\label{sec:experiments}
To understand the capabilities of \shortname{}, we evaluate each of our models on a diverse set of environments. At the same time, we try to examine our recipe for training utility models and answer a set of questions about the trained models by running a set of ablation experiments. The primary questions that we try to answer are the following:

\begin{itemize}[leftmargin=*]
    \item How well do \longname{} solve a task in an unseen environment while operating on unseen objects?
    \item What is the relative importance of different components of~\longname{}, such as training data, training algorithm, and self-verification?
          \begin{itemize}
              \item What scale of data is needed to train capable \shortname{}?
              \item What properties of data are most important for training \shortname{}?
              \item How does mLLM-based self-critique affect~\shortname{}, and where does it succeed or fail?
          \end{itemize}
    \item How well can we deploy \shortname{} on new robot embodiments?
\end{itemize}

\paragraph{Evaluation details:} For each task, we set up 25 novel environments -- five for each task -- with objects and props not seen in the training dataset. To create these evaluation environments, we take the robot to previously unseen kitchens, purchase new furniture online (door and drawer opening), and source new objects manually verified to not be in the training set (reorientation, bag and tissue pick up).
We show sample pictures of each of the environments and objects on our Appendix~\ref{sec:app:eval_envs}.
We evaluate each system and policy for 10 trials in each of these environments, starting from the same grid of starting positions facing the task space used by~\cite{shafiullah2023bringing} as we show in Appendix Figure~\ref{fig:run-schedule}.
For the retrying-based experiments, while \shortname{} take 1.31 tries in average to succeed (Section~\ref{sec:introspection-retrying}), we set a 10-try timeout to avoid getting stuck in infinite retry loops.

\subsection{Zero-shot evaluation of \shortname{} on unseen environments}
\label{sec:unseen-environments}
The most important test of capability for a \longnamesingular{} is whether such a model is capable of solving the target task in a new environment operating on new objects. We test for this capability by running our \shortname{} on our set of 25 eval environments and objects not seen during training.

\begin{figure}[ht!]
    \centering
    \includegraphics[width=\linewidth]{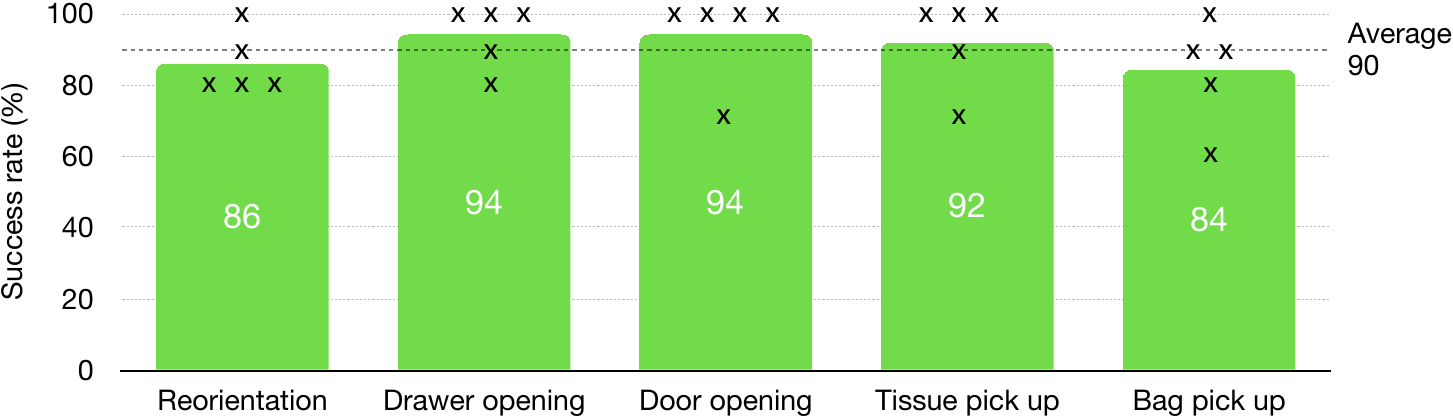}
    \caption{Success rate of~\longname{} on average over five novel scenes in five different tasks. The X's on the figure denote success rates from individual environments.}
    \label{fig:final-success-rate}
\end{figure}

On Figure~\ref{fig:final-success-rate}, we see that on unseen and novel environments, \shortname{} perform well, achieving a \finalSuccessRate{} success rate overall, and ranging between 84\% to 94\% on individual tasks. We discuss some of the failure cases we observe in the Appendix Section~\ref{app:sec:failure-modes}. Additionally, we show the performance of \shortname{} on each test environment on Table~\ref{table:success-statistics-with-retry}, showing that across all of our evaluation experiments, \shortname{} achieves some success in every environment. This success implies that our policies have a general idea of solving the target task; then such policies are further boosted with post-training methods (Section~\ref{sec:introspection-retrying}).
On all of our following experiments, we try to understand these two factors separately: the raw performance of the underlying~\shortnamesingular{} policies, and the effect of introspection and retrying on the performance of~\shortname{}.

\subsection{Effect of policy architecture and training method on \shortname{}}
\label{sec:policy-architecture}
\begin{figure}[t!]
    \centering
    \includegraphics[width=\linewidth]{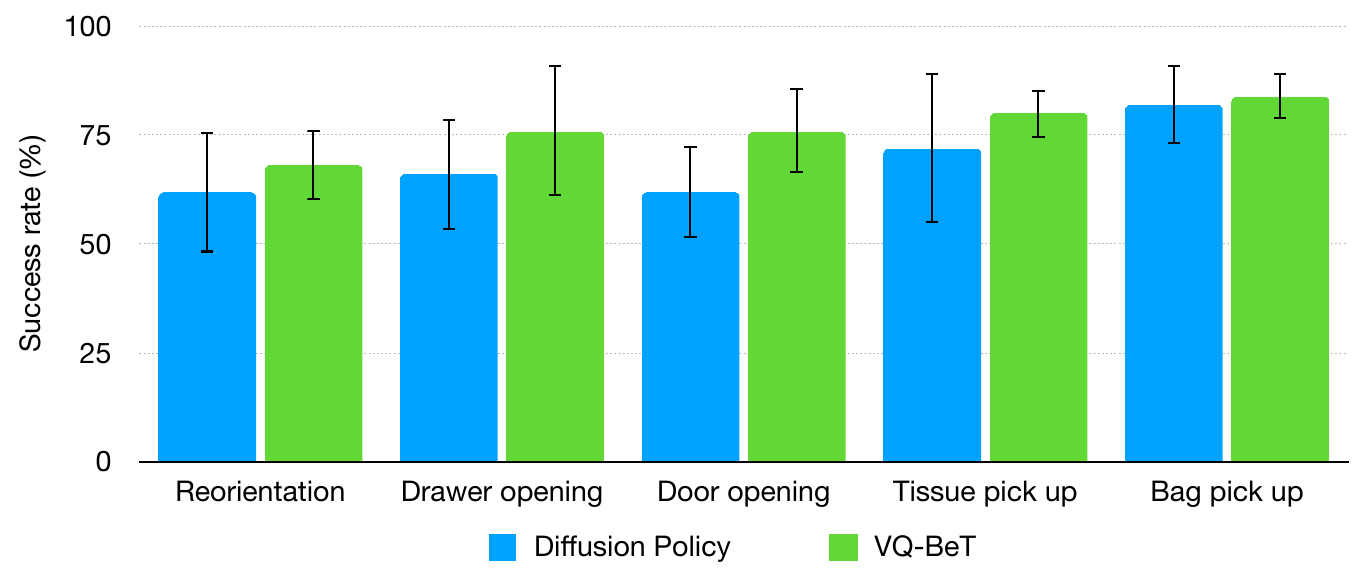}
    \caption{Relative comparison of the success rate (with standard error) of different policy architectures on our dataset on all five tasks without automated error correction. We see that the performance of VQ-BeT and Diffusion Policy is generally close, with VQ-BeT narrowly outperforming Diffusion Policy.
    }
    \label{fig:algorithm-vs-data}
\end{figure}

\begin{figure}[t!]
    \centering
    \includegraphics[width=\linewidth]{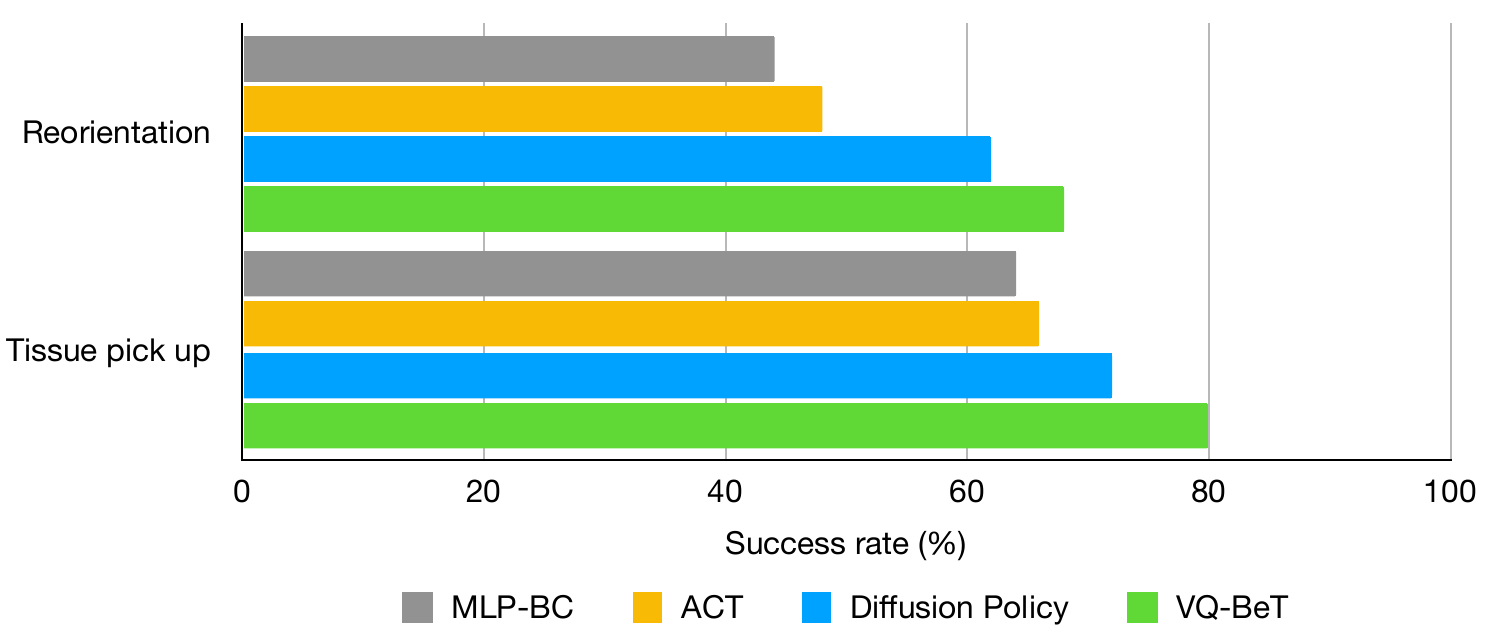}
    \caption{Relative comparison of different policy architectures on our dataset on two tasks without automated error correction. We see that while the performance of VQ-BeT and Diffusion Policy is generally neck-to-neck, while the performance of other algorithms is not far behind.
        Our experiment implies that the training data is significantly more important than training algorithm.}
    \label{fig:more-algorithm-vs-data}
\end{figure}

Once we have verified that \shortname{} can actually solve tasks in novel environments, we investigate the relative importance of different components within the training recipe. In particular, we compare the raw performance of different policy architectures on our dataset without the introspection component. We train a set of policy classes on our datasets for each task, including VQ-BeT~\citep{lee2024behavior}, Diffusion Policy (DP)~\citep{chi2023diffusion}, and as baselines, ACT~\citep{zhao2023aloha} and MLP-BC on two of the tasks. We show the relative comparison of the base success rates of different policy architectures, without retrying, in Figure~\ref{fig:algorithm-vs-data} and~\ref{fig:more-algorithm-vs-data}.

As we see in Figure~\ref{fig:algorithm-vs-data}, VQ-BeT and DP are the top two algorithms in terms of performance, with comparable performance in most tasks and overlapping error bars. Moreover, we see from Figure~\ref{fig:more-algorithm-vs-data} that while ACT and MLP-BC are not exactly on par, they are not far behind either. This observation implies that with training data of sufficient quality, the choice of algorithm may not be a make-or-break decision, and more energy should be spent on collecting diverse and accurate data. While we have similar performances on the test environment, we use VQ-BeT over DP for our final models due the higher performance and a lower latency on the robot CPU itself during deployment.
\subsection{Effect of scaling datasets on \shortname{}}
\label{sec:scaling-datasets}
\begin{figure}[t!]
    \centering
    \includegraphics[width=\linewidth]{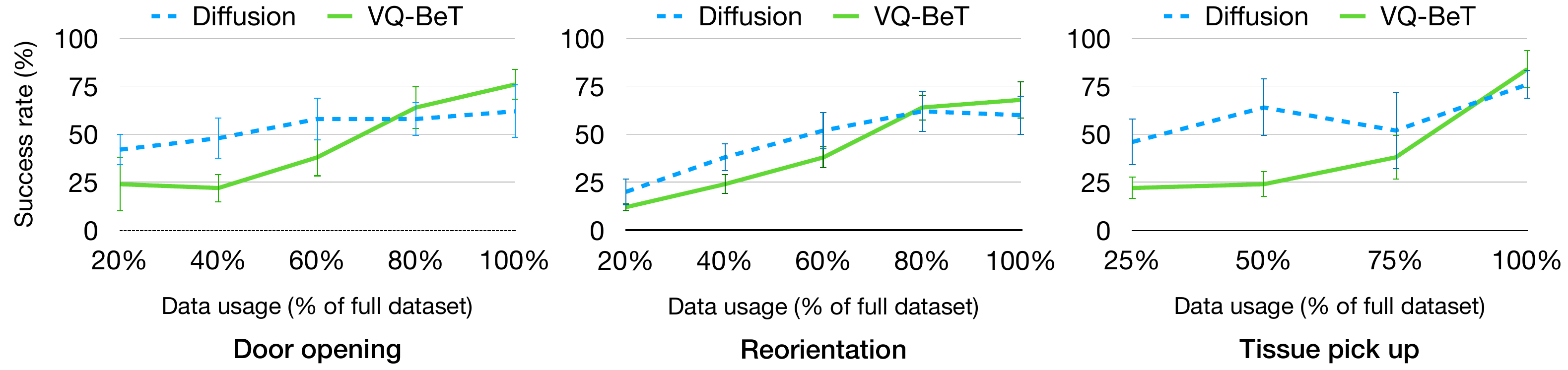}
    \caption{Understanding the performance change of \shortname{} as the dataset scales up on three of our tasks, with standard error on error bars. We see better performance from Diffusion Policy (DP) on smaller datasets, but as we scale up, VQ-BeT outperforms DP in 900--1,200 demonstrations limit.}
    \label{fig:scaling-laws}
\end{figure}
As our experiments show the importance of training data in creating~\shortname{}, we investigate the properties of the dataset that a successful~\shortname{} relies on. In particular, we dig into the scale of dataset at which reliable generalization emerges, and how \shortname' performance vary with dataset size. We train our policies on a random subset of environments from the task-specific datasets, and evaluate them on our evaluation environments.

In Figure~\ref{fig:scaling-laws}, we show the performance of VQ-BeT and Diffusion Policy without retrying trained on such data subsets on our evaluation environments as we scale up the dataset. We see that while Diffusion Policy performs better on smaller datasets, it saturates on larger datasets where VQ-BeT outperforms it. This observation implies that while a smaller dataset may be sufficient for training a capable \shortname{}, a larger dataset is crucial for achieving the best performance. Even on our largest datasets, we see that the performance of VQ-BeT continues to improve as the dataset scales up, implying that more data may improve~\shortname{} even further.
\subsection{Importance of data diversity in training \shortname{}}
\label{sec:data-diversity}
\begin{figure}[t!]
    \centering
    \includegraphics[width=\linewidth]{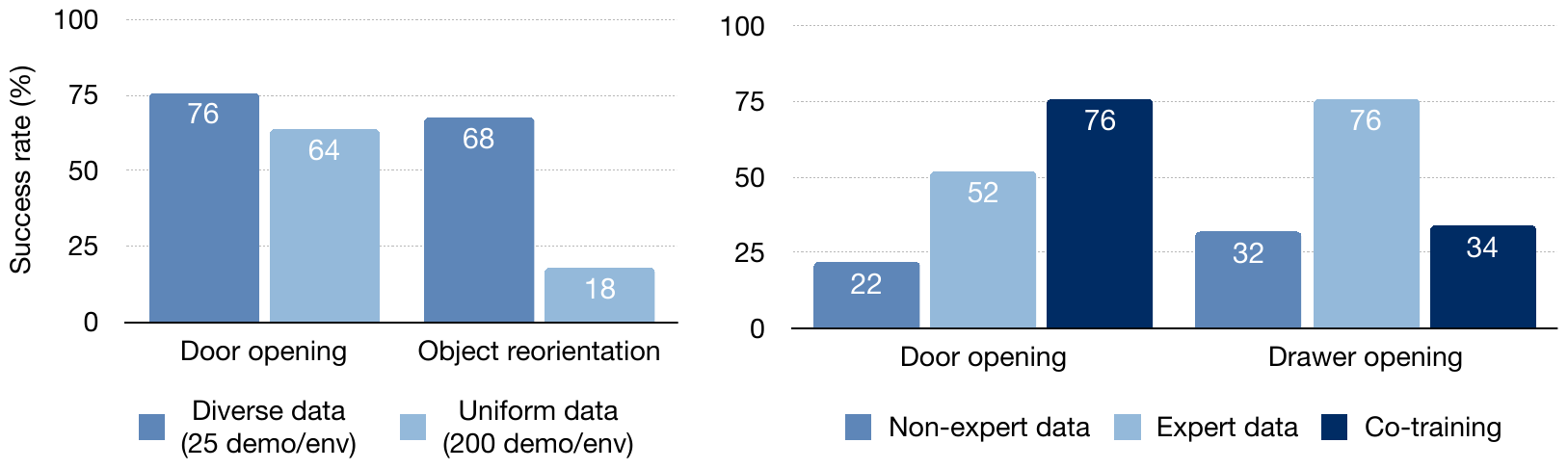}
    \caption{
        Understanding the importance of different qualities of data in training \shortname{}. On the left, we see that diverse datasets are more valuable than more uniform datasets, with strong effects on the reorientation task with many unseen environments and object.
        On the right, we see that usually expert data is more valuable than non-expert or play data while learning behavior on a same sized dataset. Moreover, we see that co-training with expert data and play data may sometimes reduce the policy performance, contrary to common knowledge.
    }
    \label{fig:expert-vs-play}
\end{figure}
Beyond the scale of the dataset, we also investigate how the diversity of the training data impacts the performance of~\shortname{} in Figure~\ref{fig:expert-vs-play} (left). We create two alternate datasets of equal size for the door opening and the object reorientation tasks. The first datasets are composed of a large number of diverse environments with roughly 25 demonstrations in each environment. The second dataset is composed of fewer, between 5 and 6, distinct environments with roughly 200 demonstrations on each environment. We see that on the door opening task, where the scene diversity is narrower, both diverse and uniform environment trained policies performed well. However, in the reorientation task, with many different unseen environments and objects, only diverse-environment trained \shortnamesingular{} policy performs well -- the policy trained on more uniform environments experiences a 50\% performance drop. This result implies that to train an effective~\shortnamesingular{}, collecting a diverse dataset is important.

\subsection{Impact of using expert demonstrations on training policies}
\label{sec:expert-vs-play}
While scaling up the dataset size and diversity is important for training~\shortname{}, an important question to consider is the quality of the training dataset. Namely, while it may be easy to collect a large number of demonstrations by a large number of demonstrators, the quality of the demonstrations may vary. In this section, we investigate the value of using expert demonstrations in training~\shortname{}.

In Figure~\ref{fig:expert-vs-play} (right) we compare the performance of~\shortname{} trained on roughly 500 demonstrations, where the data is either sampled from expert or non-expert demonstration collectors. Here, ``expertise'' is defined as experience deploying Dobb$\cdot$E policies on the robot. We see that in general, expert data is more valuable than non-expert data, with expert data outperforming non-expert data in all tasks. Moreover, we see that co-training with expert and non-expert data can sometimes, but not always, improve the performance of the policy. This observation implies depending on the task, data quality can have different levels of suboptimality, and in extreme cases may even hurt performance in co-training, which goes against a common practice in some earlier works~\citep{zhao2023aloha,khazatsky2024droid}.

\subsection{Effects of introspection and retrying with self-critique in \shortname{}}
\label{sec:introspection-retrying}
\begin{figure}[h!]
    \centering
    \includegraphics[width=\linewidth]{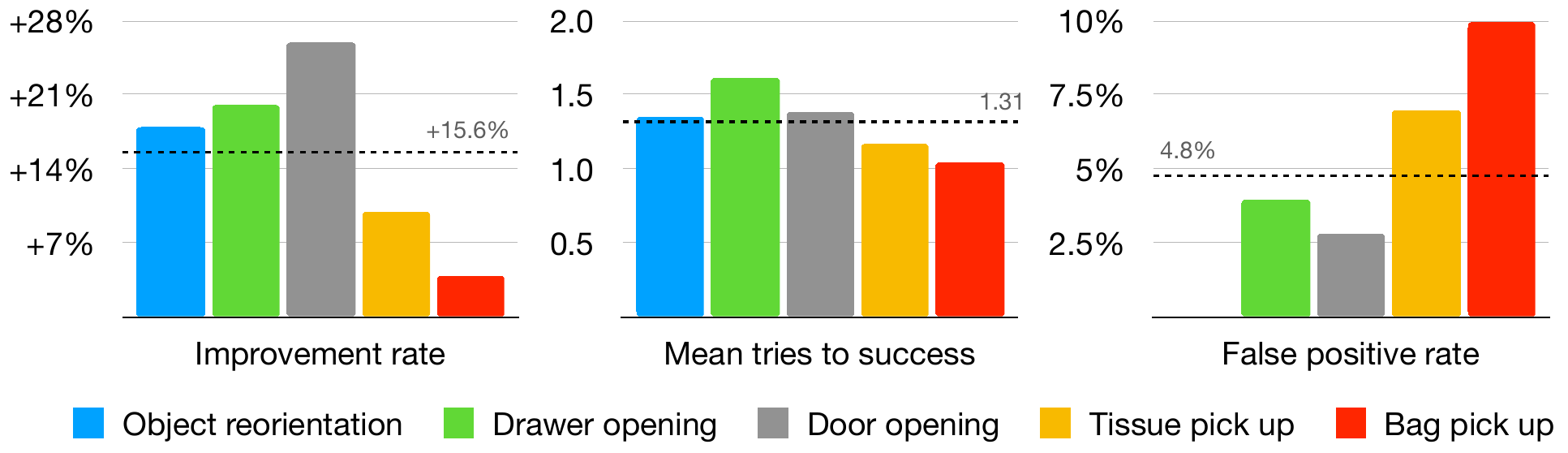}
    \caption{
        Understanding the details of introspection and retrying in \shortname{}. On the left, we see that retrying improves the performance of \shortname{} significantly, with an average 15.6\% improvement. In the middle, we see that with retrying, most tasks get solved quite fast, on average with 1.31 tries. On the right, we see that while the mLLM is able to help, it can also have false positives (4.8\% average over five tasks) which may let some errors slip past.
    }
    \label{fig:gpt-retrying}
\end{figure}
In~\shortname{}, we are using a multimodal large language model (mLLM) as a self-critique method to identify failures. However, a pretrained mLLM in practice is just another layer of fail-safe for our robot deployment, and not a guarantee of success in itself. Thus, in this section we try to understand how it helps, and how such introspection method can fail.

In Figure~\ref{fig:gpt-retrying} (left), we can see the improvement rate of using self-critique over simply using the~\shortnamesingular{} policies without any retrying mechanism. On average over our 5 tasks, we see a 15.6\% improvement over simply using~\shortnamesingular{} policies. While retrying is crucial to a higher success rate, a system that is stuck retrying for a long time is much less useful. Thankfully, on average, when~\shortname{} succeeds, it does so within 1.31 tries on average, as we see from Figure~\ref{fig:gpt-retrying} (middle). Finally, we analyze the primary failure mode of mLLMs, which is predicting false positives: classifying a trajectory as a success when it's actually a failure. On average, 4.8\% of our trajectories exhibit such behavior, constituting of half of the total errors, as seen on Figure~\ref{fig:gpt-retrying} (right).

\subsection{Transferring \shortname{} to different embodiments }
\label{sec:transferring-embodiments}
\begin{wrapfigure}{r}{0.45\textwidth}
    \vskip -0.5cm
    \centering
    \includegraphics[width=0.44\textwidth]{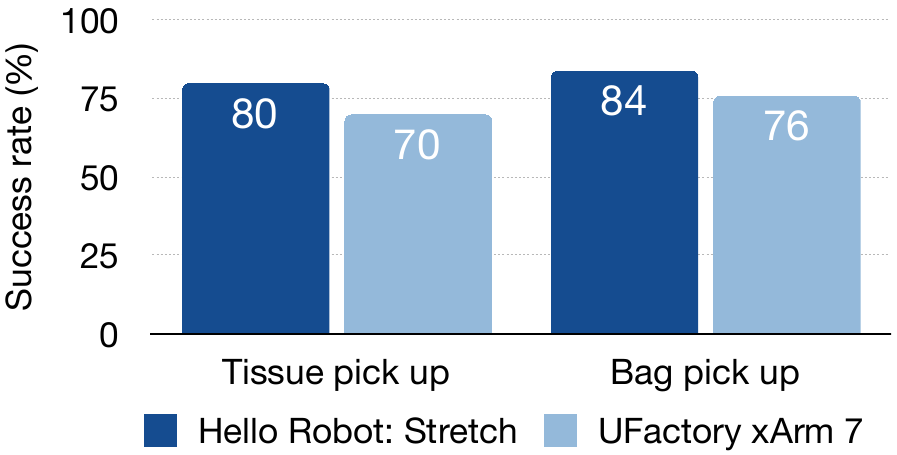}
    \caption{Performance of \shortname{} without corrections on different embodiments as shown in Figure~\ref{fig:robot-setup}: \shortname{} can transfer to different embodiments with minimal loss in performance. }
    \label{fig:embodiment-transfer}
    \vskip -0.5cm
\end{wrapfigure}

Finally, we investigate the ability of~\shortname{} to be transferred to different embodiments and cameras. We test the performance of two~\shortname{} on the other robot setup shown in Figure~\ref{fig:robot-setup}: UFactory xArm 7, which is different from the Hello Robot Stretch setup we run other experiments on. We see that~\shortname{} can be transferred to different embodiments and cameras with minimal loss in performance: roughly 10\% drop in performance in both cases without corrective mLLM feedback, as shown in Figure~\ref{fig:embodiment-transfer}. We expect combining~\shortname{} with the mLLM self-critique would result in similar increase in performance in other embodiments as well; in fact, with an external third person camera, we expect to see a higher portion of the errors being caught and corrected. This experiment implies that~\shortname{} can be easily deployed on different robots and cameras with minimal effort, making it a versatile tool for a wide range of robotic applications.

\section{Related works}
\label{sec:related_work}

\paragraph{Large Scale Data Collection:} The data acquisition pipeline represents one of the most critical element of a data-driven robot learning framework. Previous works has employed a diverse array of data acquisition techniques, combining many open-sourced datasets across diverse simulation or real-world data including diverse robot embodiment from many institutions across the globe ~\citep{reed2022a, padalkar2023open, brohan2023rt2, khazatsky2024droid}.

The most common approaches to robot demonstration collection involves pairing the robot or end-effector with remote controller devices or kinematically isomorphic equipment. The devices utilized have a range of complexity and forms: they encompass full robotic exoskeletons ~\citep{zhao2023wearable, ishiguro2020bilateral, fang2023airexo}, as well as simpler data collection tools ~\citep{zhao2023aloha, wu2023gello,fu2024mobile}, and also methods that don't require physically moving a robot ~\citep{shafiullah2023bringing, song2020grasping, pari2021surprising, young2020visual, chi2024universal}. Additionally, various control methods have also been employed, including the use of video game controllers ~\citep{liu2024libero, sian2004whole}, Virtual Reality (VR) devices ~\citep{iyer2024open, cui2022play, cheng2024opentelevisionteleoperationimmersiveactive, yang2024acecrossplatformvisualexoskeletonslowcost,park2024using,arunachalam2022dexterous,arunachalam2023holodex,fu2024humanplus}, and mobile phones~\citep{mandlekar2018roboturk}.

While the most intuitive method is to physically move a real robot, it is both difficult to do and hard to scale to a diverse set of environments. The hardware controller approach can be inefficient because it requires the demonstrator to mentally map robot behavior to controller inputs. The opposite, using a device without moving the robot is efficient in that the demonstrator's movements can be mapped directly to the robot, but it is challenging to apply force feedback. Studies that provides perspective on the relative merits of these two direction are~\citep{shafiullah2023bringing, chi2024universal}, which combines the versatility of simple controller with the intuitiveness of moving a physical end-effector. In this work, we employ a device that inherits and improves the device proposed from ~\citep{shafiullah2023bringing, chi2024universal} for our data collection pipeline.

\paragraph{Pretrained Robot Models:}
Pre-trained foundation models have demonstrated a wide range of generalization performance across various domains, with the capability to learn from internet-scale pre-training data ~\citep{devlin2018bert, radford2021clip, dubey2024llama, kirillov2023segment}. However, in comparison to these vision and language pre-trained models, learning a foundation model for robotics has been considered a relatively challenging area, due to the limited quantity of available datasets ~\citep{kappler2015leveraging, levine2016end, depierre2018jacquard, zhu2023fanuc}, the significant discrepancy across the domains~\citep{dasari2019robonet, kalashnikov2021mt, padalkar2023open}, and the inherently challenging nature of the action datasets in terms of tokenization~\citep{lee2024behavior, brohan2023rt2, zhengprise}.

To address these issues, recent research is increasingly adopting techniques that introduce modular and hierarchical systems, incorporate pre-trained language and visual models~\citep{li2023vision, nair2022r3m, karamcheti2023language, shafiullah2022clip, liu2024ok, gupta2024opening}, and collect large scale data with efficient data collection schemes~\citep{khazatsky2024droid, brohan2023rt2, walke2023bridgedata, ebert2021bridge, fang2023rh20t}. Consequently, they have enabled the pre-trained foundation robot models to exhibit enhanced generalization performance, thereby showcasing that the robotic agents are capable of operating in more than one robot embodiment and operating environment~\citep{team2024octo, reed2022a, kim2024openvla, doshi2024scaling}. In contrast with the aforementioned approaches, which follow a method of training on internet-scale data and fine-tuning on task-specific data, our approach does not expect that the model will have access to a dataset in the environments where the robot is expected to operate. Rather, this project demonstrates the capacity of generalizable performance without a necessity to fine-tune the model for each novel robot embodiment and environment.

\paragraph{Large Models Feedback and Improvement:} Due to their capacity to comprehend intricate semantics and relations, Natural language and Large language models (LLM), have recently been applied to robotic agents powered by imitation learning ~\citep{fried2018speaker-follower, kim2024openvla, shridhar2022cliport, jang2022bc} and reinforcement learning~\citep{du2023guiding, goyal2021pixl2r}.

Among the wide capabilities afforded by language models, those commonly employed in the context of decision-making include providing feedback in the resolution of uncertain information ~\citep{ren2023robots, mullen2024towards, huang2022inner,liu2023reflect,guo2023doremi,park2023clara,gao2024physically}, suggesting affordance of what is possible in the environments by combining with Value functions~\citep{ahn2022can}, and imagination of outcomes \citep{zhang2024rail} or planning and decompose complex tasks into mid-level plans~\citep{song2023llm, huang2022language, zeng2022socratic, sharma2021skill}. Language models could also be used to improve the overall performance of autonomous agent systems by improving reward signal ~\citep{nair2022learning, goyal2021pixl2r, ma2023eureka}, leveraging their long-horizon reasoning ~\citep{dalal2024plan, zhou2023generalizable, blukis2022persistent}, or designing environments ~\citep{ma2024dreureka}. In this project, we employ the mLLM to provide feedback in the form of a reset signal in open-ended environments, a manner analogous to that of the studies above.
\section{Limitations and Conclusion}
While in this work we create~\longname{} that can perform particular tasks zero-shot in novel environments, there are certain limitations that future versions can improve upon. The primary limitation that we see are of hardware: for example, two-fingered grippers like our Stick-v2 are unable to open doors with round doorknobs. Similarly, while flexible fingertips can be more lenient for the policy, it makes it hard to manipulate heavy objects. We encourage more research on better gripper and fingertip design to address these issues. Secondly, we assume navigation to be a separate component, and in this work assume that the robot is in the task space facing the task objective. Combining with modular navigation work such as~\citep{liu2024ok} should address this issue. Finally, for mLLM introspection and retrying, we assume that the errors made by our model (a) leaves the task-space somewhat in-distribution, and (b) allows for an easy reset of the robot to the initial state. Increasing training data with failure recovery behavior in our dataset should let our robots recover more naturally from such failure cases.

\section*{Acknowledgements}
We thank Shenglong Wang and the NYU HPC team for helping us with compute, Blaine Matulevic and Binit Shah for supporting our hardware needs, and Siddhant Haldar and Jeff Cui for providing feedback on the paper. NYU authors are supported by grants from Honda, Hyundai, NSF award 2339096 and ONR awards N00014-21-1-2758 and N00014-22-1-2773. MS is supported by the Apple Fellowship. LP is supported by the Packard Fellowship. SL is supported by the Daishin Songchon Foundation. Hello Robot authors are supported by NIH NIA R43AG072982.

\newpage
\bibliography{references}

\newpage
\appendix
\section{Appendix}
\subsection{Detailed Results from Experiments with Self-critique and Retrying}
\label{sec:app:auto-retrying}

\begin{table}[ht]
\centering
\captionsetup{justification=centering}
\begin{tabular}{llr}
\toprule
\textbf{Task} & \textbf{Environment/Object} & \textbf{Success $\cdot / 10$} \\
\midrule
\textbf{Door Opening} & Kitchen Trash Door & 7 \\
                      & Kitchen Cabinet Door & 10 \\
                      & Brown Cabinet Door & 10 \\
                      & Metal Cabinet Door & 10 \\
                      & White File Cabinet Door & 10 \\
\midrule
\textbf{Drawer Opening} & Kitchen Drawer & 10 \\
                        & Cloth Drawer & 9 \\
                        & White File Cabinet Drawer & 10 \\
                        & Small File Cabinet Drawer & 10 \\
                        & Dresser Drawer & 8 \\
\midrule
\textbf{Bag Pick Up} & Hollister Bag & 9 \\
                     & American Eagle Bag & 10 \\
                     & Qdoba Bag & 8 \\
                     & Journey's Bag & 9 \\
                     & Yellow Bag & 6 \\
\midrule
\textbf{Tissue Pick Up} & White Tall Box & 10 \\
                        & White Short Box & 10 \\
                        & Black Square Box & 9 \\
                        & Red Square Box & 10 \\
                        & Kleenex Box & 7 \\
\midrule
\textbf{Object Reorientation} & Pink Bottle & 9 \\
                              & White Board Cleaner & 8 \\
                              & Spices Container & 8 \\
                              & Coke Can & 8 \\
                              & Compressed Air & 10 \\
\bottomrule
\end{tabular}
\caption{Detailed success statistics of~\shortname{} on our evaluation environments.}
\label{table:success-statistics-with-retry}
\end{table}
\newpage
\subsection{Evaluation Environments}
\label{sec:app:eval_envs}

\begin{figure}[ht!]
    \centering
    \includegraphics[width=0.9\linewidth]{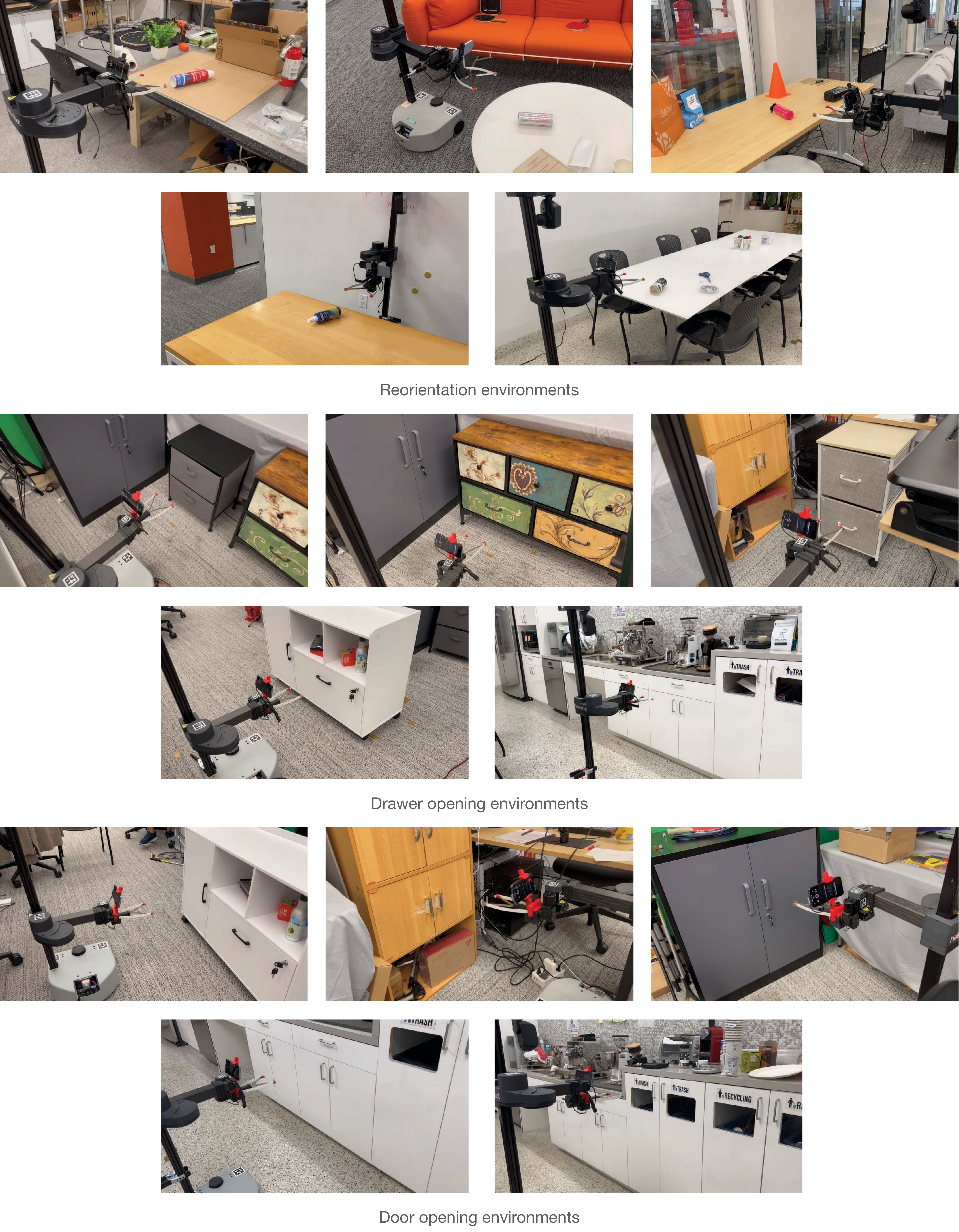}
    \caption{Picture of evaluation environments for the tasks Reorientation, Drawer opening, and Door opening.}
    \label{fig:eval_envs_1}
\end{figure}

\begin{figure}[ht!]
    \centering
    \includegraphics[width=0.9\linewidth]{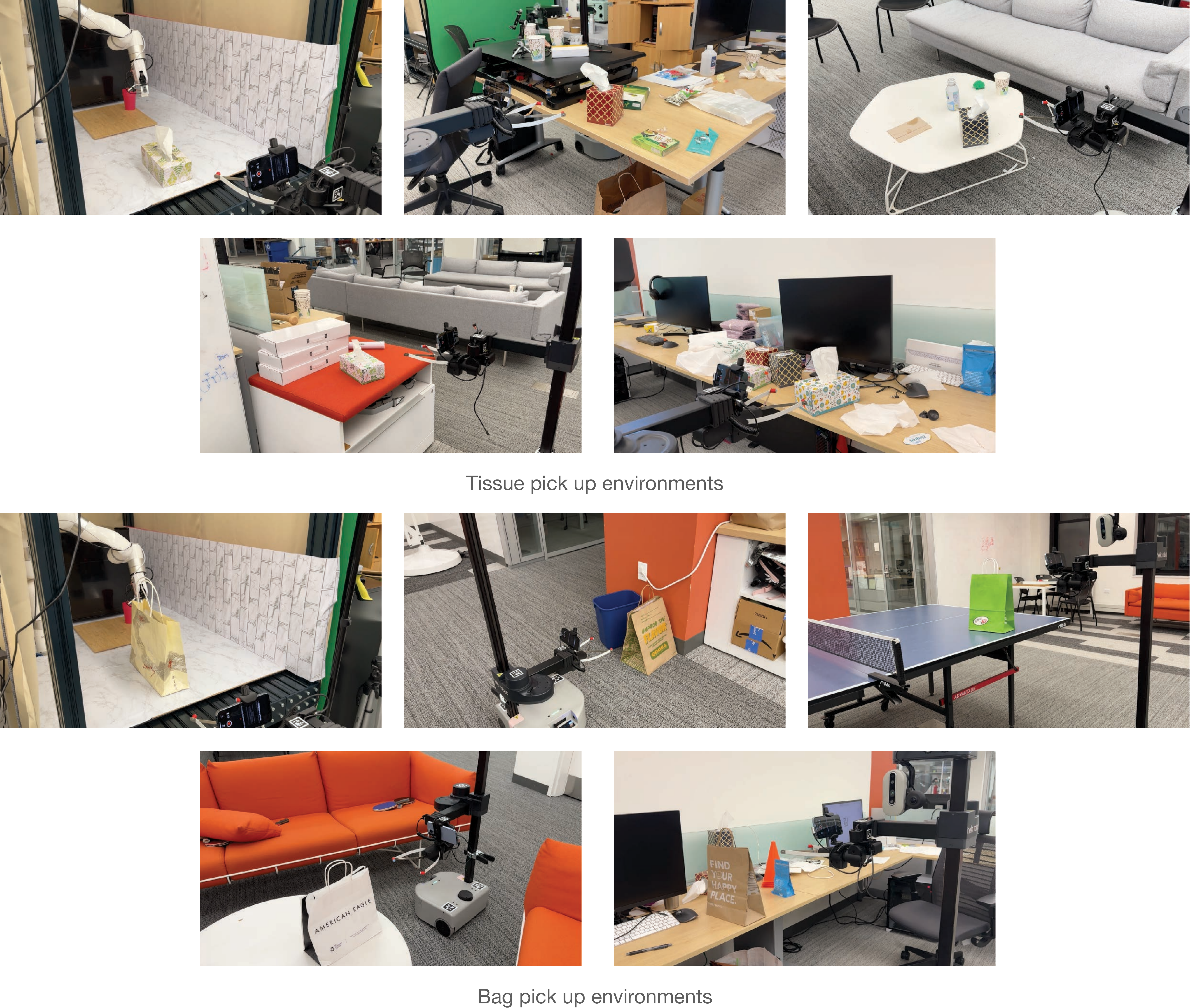}
    \caption{Pictures of the evaluation environments for the task Tissue pick up and Bag pick up.}
    \label{fig:eval_envs_2}
\end{figure}
\newpage

\subsection{Multimodal Large Language Model Prompts for Success Verification}
\label{sec:app:mllm-prompt}
Here, we present the prompt that we use to verify~\shortname{} success with mLLMs.
\definecolor{violetbg}{RGB}{176, 88, 228}
\definecolor{violetframe}{RGB}{112, 41, 156}
\definecolor{white}{RGB}{255, 255, 255}

\tcbset{
  myboxstyle/.style={
    colback=violetbg!20,
    colframe=violetframe,
    width=0.8\textwidth, 
    arc=5mm,
    boxrule=0.5mm,
    left=5mm,
    right=5mm,
    top=2mm,
    bottom=2mm,
    fonttitle=\bfseries,
    coltitle=white
  }
}

\begin{center}
\begin{tcolorbox}[title=Door Opening, myboxstyle]
As the timesteps progress, does the robotic arm open the door AND is the robot arm grasping the handle in the LAST timestep?\\
\textbf{Please respond with only 'Yes' or 'No'.}
\end{tcolorbox}

\vspace{5mm}

\begin{tcolorbox}[title=Drawer Opening, myboxstyle]
As the timesteps progress, does the robotic arm grasp the drawer handle and open it AND is the drawer open in the last timestep?\\
\textbf{Please respond with only 'Yes' or 'No'.}
\end{tcolorbox}

\vspace{5mm}

\begin{tcolorbox}[title=Reorientation, myboxstyle]
As the timesteps progress, does the robotic arm/gripper reorient the object upright AND is the object upright in the LAST frame?\\
\textbf{Please respond with only 'Yes' or 'No'.}
\end{tcolorbox}

\vspace{5mm}

\begin{tcolorbox}[title=Tissue Pick-Up, myboxstyle]
As the timesteps progress, does the robotic arm/gripper grasp the tissue AND is the gripper grasping the tissue in the LAST timestep?\\
\textbf{Please respond with only 'Yes' or 'No'.}
\end{tcolorbox}

\vspace{5mm}

\begin{tcolorbox}[title=Bag Pick-Up, myboxstyle]
As the timesteps progress, does the robotic arm/gripper grasp the bag AND is the gripper grasping the bag in the LAST timestep?\\
\textbf{Please respond with only 'Yes' or 'No'.}
\end{tcolorbox}
\end{center}

\subsection{Evaluation Schedule}
\begin{figure}[h!]
    \centering
    \includegraphics[width=0.75\linewidth]{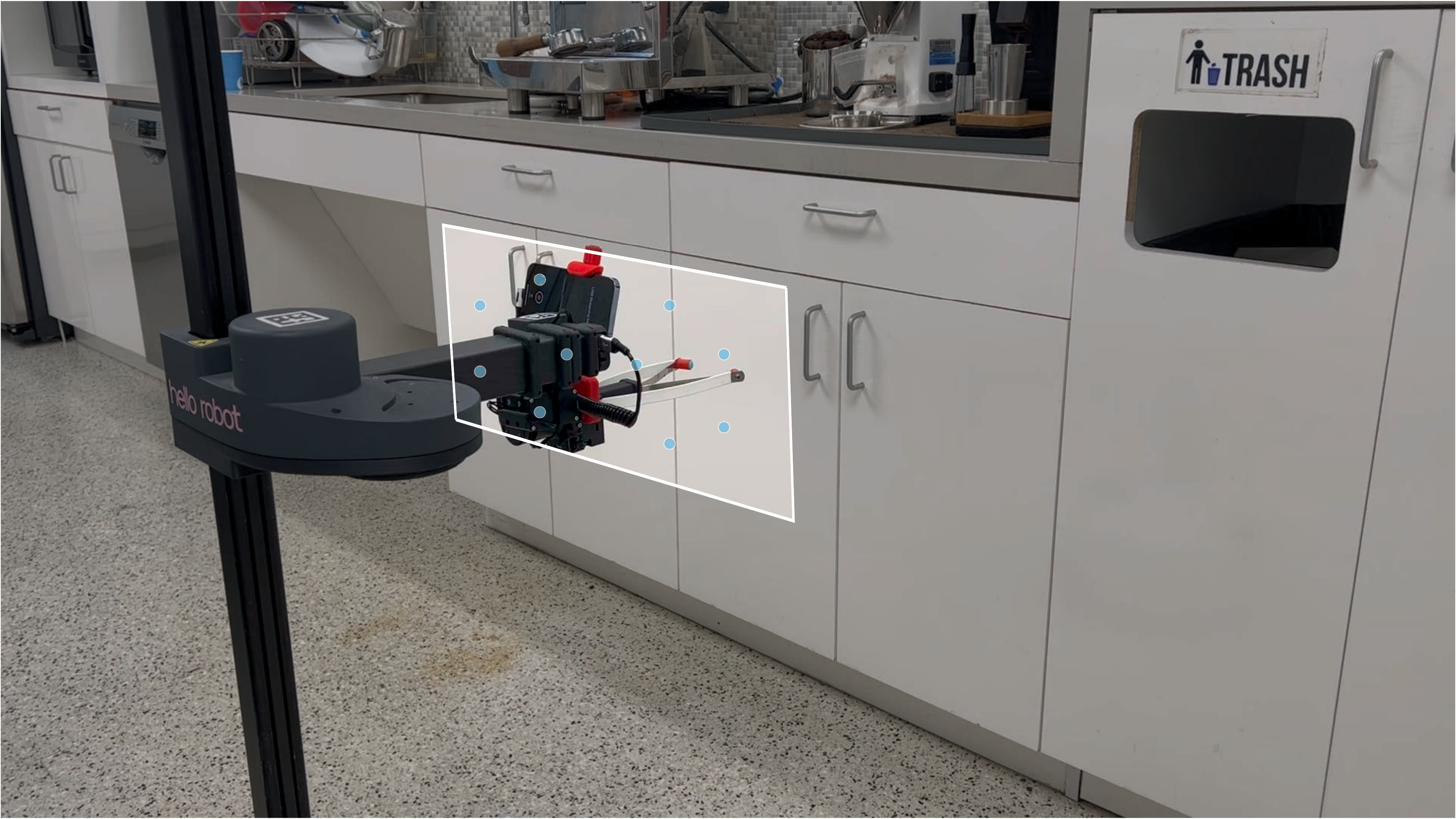}
    \caption{10-run evaluation schedule used to evaluate \longname{}, with robot starting positions denoted by the pale blue dots in the image. We assume that the robot is at the task space facing the object, but it can be at different offsets with respect to the target object. On our object centric tasks (reorientation, bag and tissue pickup) we also randomize the position of the object itself.}
    \label{fig:run-schedule}
\end{figure}

In Figure~\ref{fig:run-schedule}, we show the starting position of the robot for our 10-run evaluations to understand the positional generalization capabilities of \longname{}.

\subsection{Bill of Materials}
\label{sec:app:bom}
Here, we present the bill of materials for our hardware components, assuming that the interested researcher or user owns an iPhone Pro already. The total cost comes out to be slightly below $\$25$ for the entire setup.

\begin{table}[ht]
\centering
\captionsetup{justification=centering}
\begin{tabular}{lrrr}
\toprule
\textbf{Item} & \textbf{Price} & \textbf{Unit Price} & \textbf{Qty} \\
\midrule
Reacher Grabber Tool & 26.99 & 13.50 & 1 \\
Brass Tapered Heat-Set Inserts & 21.82 & 0.22 & 3 \\
Thread-Forming Screws & 7.75 & 0.31 & 3 \\
Button Head Screw - M4 x 0.70 - 8mm & 12.91 & 0.13 & 1 \\
Button Head Screw - M4 x 0.70 - 5mm & 8.64 & 0.09 & 2 \\
Button Head Screw - M4 x 0.70 - 35mm & 16.77 & 0.34 & 2 \\
Nylon-Insert Locknut & 5.57 & 0.06 & 2 \\
Dowel Pin & 16.09 & 0.32 & 3 \\
Nylon Unthreaded Spacer & 18.41 & 0.18 & 2 \\
Kevlar Cord & 20.99 & 20.99 & 1/100 \\
Heat Shrink Tubing & 10.79 & 10.79 & 1/30 \\
Black 3D Printer Filament & 25.99 & 25.99 & 3/20 \\
\midrule
\textbf{Total} & & {\textbf{21.99}} \\
\bottomrule
\end{tabular}
\caption{Stick-v2 Main Body}
\end{table}

\begin{table}[ht]
\centering
\captionsetup{justification=centering}
\begin{tabular}{lrrr}
\toprule
\textbf{Item} & \textbf{Price} & \textbf{Unit Price} & \textbf{Qty} \\
\midrule
Socket Head Screw - M3 x 0.5mm - 8mm & 12.52 & 0.13 & 2 \\
Steel Hex Nut - M3 x 0.5mm & 2.62 & 0.03 & 2 \\
M3 Steel Washer & 2.19 & 0.02 & 2 \\
Red 3D Printer Filament & 25.99 & 25.99 & 3/1000 \\
Oomoo 25 Silicone Rubber & 33.99 & 33.99 & 1/200 \\
\midrule
\textbf{Total} & & {\textbf{0.61}} \\
\bottomrule
\end{tabular}
\caption{Gripper Tips}
\end{table}

\begin{table}[ht]
\centering
\captionsetup{justification=centering}
\begin{tabular}{lrrr}
\toprule
\textbf{Item} & \textbf{Price} & \textbf{Unit Price} & \textbf{Qty} \\
\midrule
Socket Head Screw - M5 x 0.8mm - 20mm & 17.10 & 0.17 & 1 \\
Socket Head Screw - M5 x 0.8mm - 50mm & 4.26 & 0.85 & 1 \\
Steel Hex Nut - M5 x 0.8mm & 5.24 & 0.05 & 2 \\
Button Head Screw - M4 x 0.70 - 8mm & 12.91 & 0.13 & 1 \\
Black 3D Printer Filament & 25.99 & 25.99 & 3/20 \\
\midrule
\textbf{Total} & & {\textbf{2.03}} \\
\bottomrule
\end{tabular}
\caption{Phone Holder}
\end{table}
\newpage
\subsection{Deploying on Stretch's Default D405 Camera}
\label{app:sec:d405}
Deploying our~\longname{} on the standard Hello Robot Stretch SE3 requires normalizing the image coming out of the default Intel Realsense D405 wrist camera. We created an affine transformation that maps the D405 image to the same pixel coordinates as the iPhone camera.
\begin{figure}[h!]
    \centering
    \includegraphics[width=\linewidth]{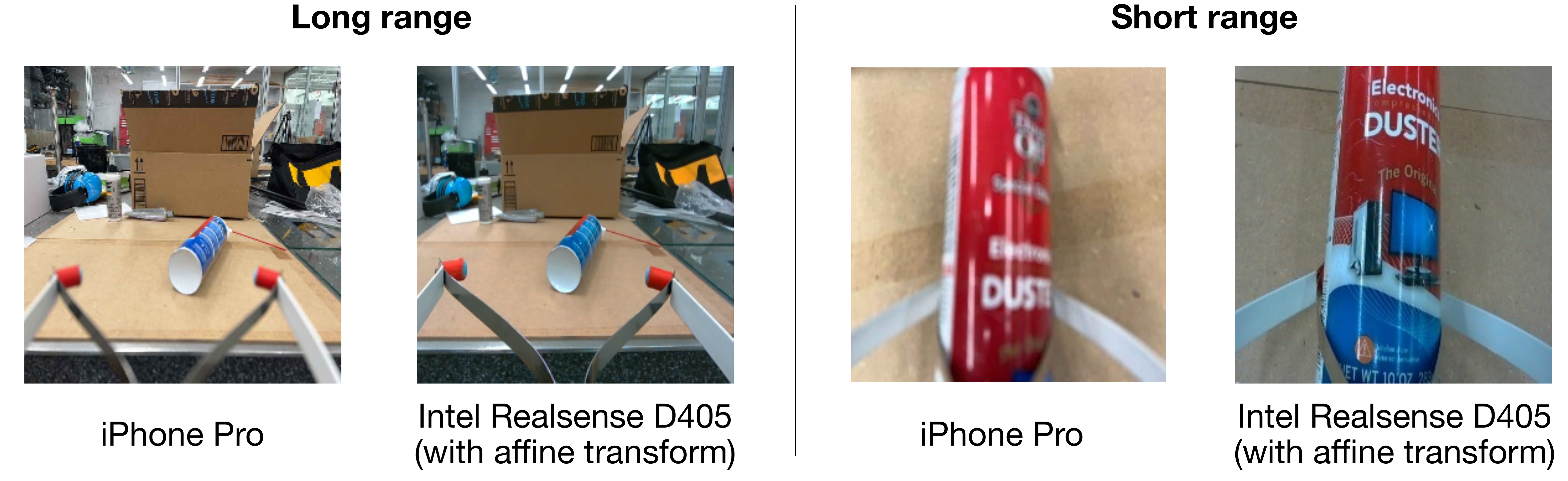}
    \caption{We can see the corresponding D405 camera image alongside the iPhone Pro image. While in the long range, the images look similar, in the short range iPhone images are out of focus because of the different focal lengths of the cameras.}
    \label{fig:d405-transform}
\end{figure}

As we can see from Figure~\ref{fig:d405-transform}, applying the affine transform to the D405 camera maps it to pretty similar viewpoint as the wrist mounted iPhone. While we can run~\shortname{} directly with this camera transform, we see a performance drop which we hypothesize happens because of the especially apparent difference in close-range. This difference is caused by the different focal lengths of the two cameras, and may be solved in the future with image augmentations.
\newpage

\subsection{Failure Modes}
\label{app:sec:failure-modes}
\begin{figure}[h]
    \centering
    \includegraphics[width=\linewidth]{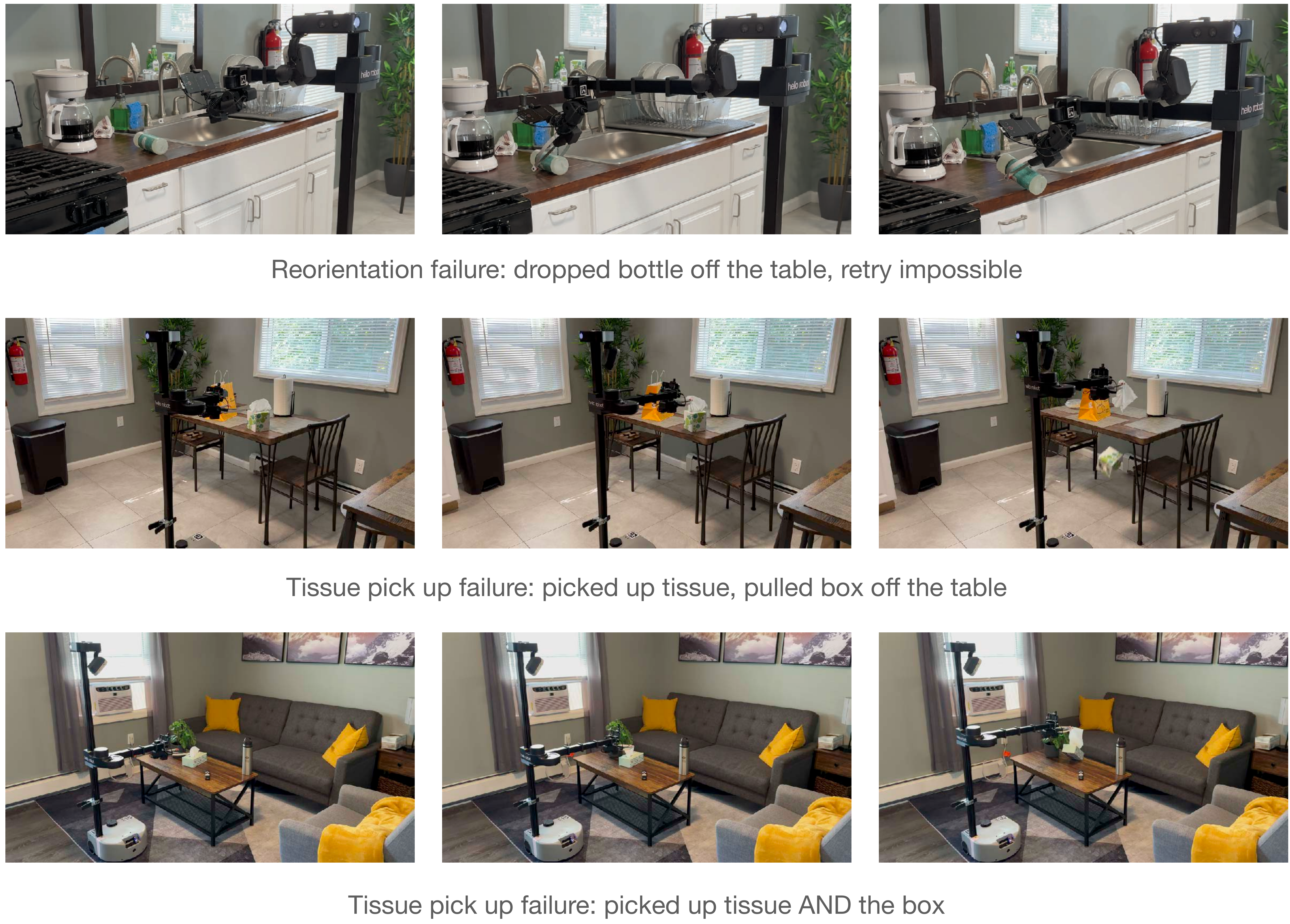}
    \caption{Examples of some failures in real world rollouts. Since~\shortname{} retries on failure with mLLM feedback, the failure modes tend to be peculiar, some examples of which are shown here.}
    \label{fig:failure-examples}
\end{figure}

As we mention in the main paper, with mLLM guided retries, our failures tend to be more peculiar than simply ``robot failed to complete task''. In this section, we try to shine some light on what kind of failures we experience in our system.

\begin{itemize}[leftmargin=*]
    \item \textbf{Reorientation:} Primary failure modes for this task are when retry becomes impossible because of environmental issues, such as the target bottle rolling away on the table, being dropped off the surface (an example of which is shown on the Figure~\ref{fig:failure-examples}), pushing it too far into the table (to a position too far for our robot arm), or being rotated sideways by the gripper before grasping. In out-of-distribution surfaces, it can be hard to estimate how large the surface is visually and thus placing the object after reorientation may miss the surface or the robot may run into the surface.
    \item \textbf{Drawer opening:} Beyond the most direct failure mode of missing the drawer handle, we experienced some failure modes where the model does not know when to stop pulling on cloth drawers and thus pulls out the entire drawer. Without force feedback, it can be hard to tell visually when the drawer starts sagging. Force feedback on the fingertips would help the robot correct for it.
    \item \textbf{Door opening:} Here, the primary failure mode we experience are on unusual doors, such as the trash cabinet door with a hole in it. There, GPT sometimes classifies the door as ``open'' even when it is closed. In some rare cases, when door handles are close together, the robot may grasp around both handles and fail to reset as it gets stuck when retracting.  
    \item \textbf{Tissue pick up:} The tissue box itself being light and easy to move means that sometimes the box moves with the tissue as its being picked up. As a result, the box may get picked up with the tissue, or get pushed off from its table by the robot (Figure~\ref{fig:failure-examples}.)
    \item \textbf{Bag pick up:} The case of bag picking up is interesting because it has one of the highest success rates from the raw~\shortnamesingular{} policy but also sees the smallest improvement (4\%) from GPT feedback. This failure from mLLM feedback happens usually because from the robot wrist or head camera, it can be hard to tell whether the bag has been picked up. As a result, GPT tends to have a high number of false positives for this task. Having a better third-person view of the workspace should help address this issue.
\end{itemize}

\end{document}